\documentclass[sigconf]{acmart}
\usepackage{multicol}
\usepackage{seqsplit}
\usepackage{graphicx}
\usepackage{xcolor}
\usepackage{multirow}
\usepackage{booktabs}
\usepackage{colortbl}

\usepackage[frozencache=true,cachedir=minted-cache]{minted}
\definecolor{brickred}{RGB}{170, 74, 68}

\AtBeginDocument{%
  \providecommand\BibTeX{{%
    \normalfont B\kern-0.5em{\scshape i\kern-0.25em b}\kern-0.8em\TeX}}}

\setcopyright{acmcopyright}

\copyrightyear{2023}
\acmYear{2023}
\setcopyright{acmlicensed}\acmConference[HuMob-Challenge '23]{1st International Workshop on the Human Mobility Prediction Challenge}{November 13, 2023}{Hamburg, Germany}
\acmBooktitle{1st International Workshop on the Human Mobility Prediction Challenge (HuMob-Challenge '23), November 13, 2023, Hamburg, Germany}
\acmPrice{15.00}
\acmDOI{10.1145/3615894.3628499}
\acmISBN{979-8-4007-0356-0/23/11}

\begin{document}

\title[GeoFormer: Predicting Human Mobility using GPT]{GeoFormer: Predicting Human Mobility using Generative Pre-trained Transformer (GPT)}


\author{Aivin V. Solatorio}
\authornote{The findings, interpretations, and conclusions expressed in this paper are entirely those of the authors. They do not necessarily represent the views of the International Bank for Reconstruction and Development/World Bank and its affiliated organizations, or those of the Executive Directors of the World Bank or the governments they represent. GitHub/HuggingFace: @avsolatorio. \url{https://github.com/avsolatorio/geoformer}}
\email{asolatorio@worldbank.org}

\affiliation{%
  \institution{The World Bank}
  \streetaddress{1818 H Street NW}
  \city{Washington}
  \state{District of Columbia}
  \country{USA}
  \postcode{20433}
}

\renewcommand{\shortauthors}{Solatorio, Aivin V.}

\begin{abstract}

    Predicting human mobility holds significant practical value, with applications ranging from enhancing disaster risk planning to simulating epidemic spread. In this paper, we present the GeoFormer, a decoder-only transformer model adapted from the GPT architecture to forecast human mobility. Our proposed model is rigorously tested in the context of the HuMob Challenge 2023—a competition designed to evaluate the performance of prediction models on standardized datasets to predict human mobility. The challenge leverages two datasets encompassing urban-scale data of 25,000 and 100,000 individuals over a longitudinal period of 75 days. GeoFormer stands out as a top performer in the competition, securing a place in the top-3 ranking. Its success is underscored by performing well on both performance metrics chosen for the competition—the GEO-BLEU and the Dynamic Time Warping (DTW) measures. The performance of the GeoFormer on the HuMob Challenge 2023 underscores its potential to make substantial contributions to the field of human mobility prediction, with far-reaching implications for disaster preparedness, epidemic control, and beyond.

\end{abstract}







\maketitle

\section{Introduction}


Digitalization has unlocked a tremendous amount of data useful for assessing human mobility captured through call detail records (CDRs) or GPS-enabled devices and smartphones. Human mobility has become increasingly impactful in today's world either as a direct or proxy indicator for various socioeconomic activities, including as input to other applications ranging from disease spread modeling during the COVID-19 pandemic to influencing transport and urban planning decisions. Accurate forecasts of human mobility patterns can empower policymakers, urban planners, and healthcare professionals with valuable insights to better prepare for various scenarios.

The HuMob Challenge 2023 aims to find mobility prediction models that can predict individual human trajectories contained in two standardized benchmarking datasets \cite{noauthor_humob_nodate}. The datasets encompass two distinct types of human mobility: normal period mobility and emergency period mobility, each holding unique challenges and implications. In response to the challenge, we introduce a novel approach to predicting human mobility that leverages a decoder-only transformer model, specifically the generative pre-trained transformer (GPT) architecture \cite{radford2018improving}.


The recent popularity of artificial intelligence (AI) applications has largely been driven by the phenomenal and almost human-like capability of generative deep learning models. The release of ChatGPT, a conversational application of generative AI by OpenAI, is arguably the inflection point for the mainstream adoption of AI. ChatGPT, at its core, is powered by transformers \cite{vaswani2017attention}—a revolutionary and now a ubiquitous component for a large number of state-of-the-art models across use cases \cite{solatorio2023realtabformer}. With the goal of introducing generative AI in human mobility prediction, this work makes three contributions to the field. Firstly, we present a pioneering approach that offers an innovative perspective on modeling and predicting human mobility using generative deep learning models. Secondly, we further demonstrate the versatility and cross-functional applicability of generative models built on transformer architectures, establishing their potential as a fundamental tool in the domain of mobility prediction. Lastly, we layout the process of integrating insights from mobility data into transforming the human mobility problem into a problem that is analogous to those in natural language processing (NLP).

\begin{figure*}[h]
  \centering
  \includegraphics[width=\textwidth]{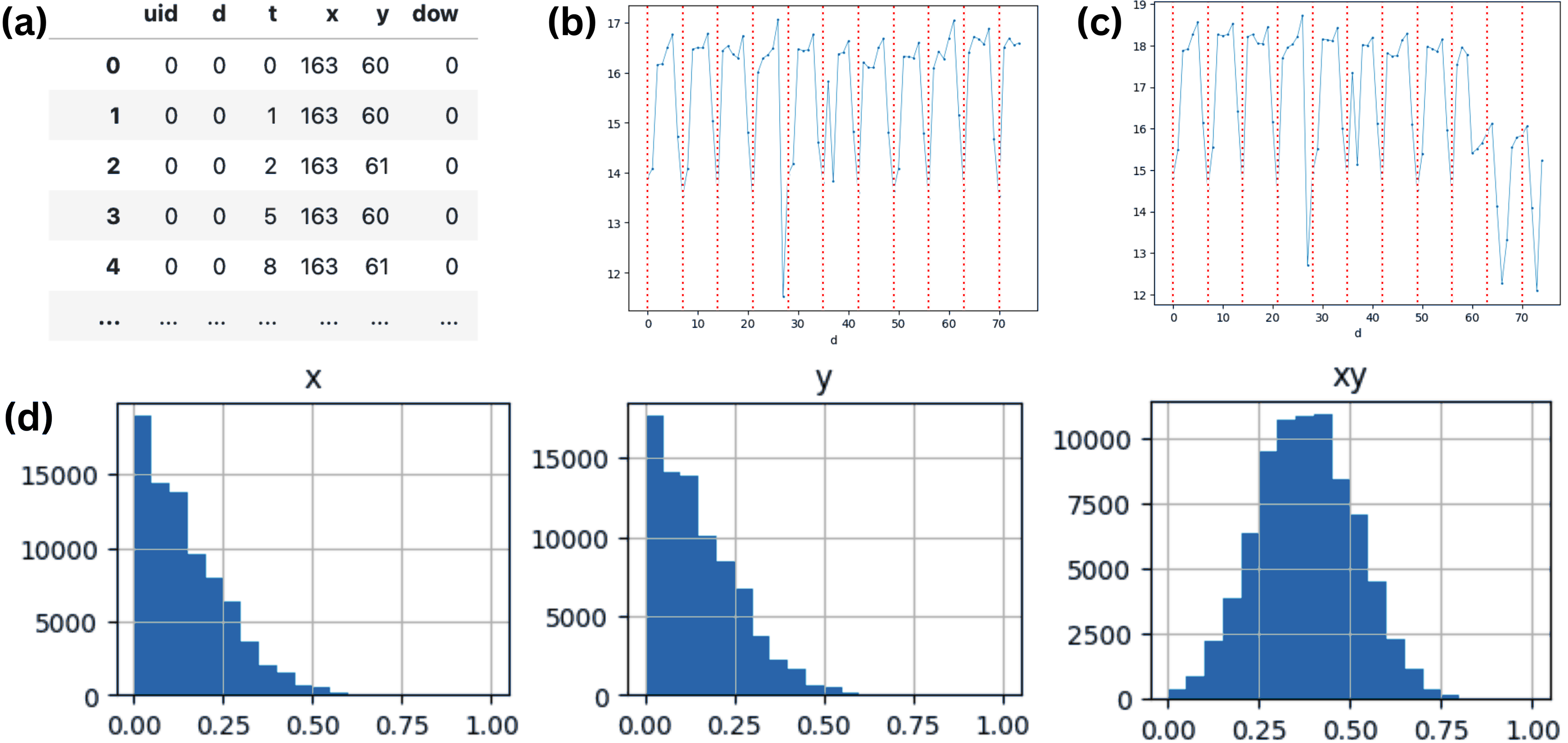}
  \caption{Panel (a) shows the snapshot of the raw data structure including the day-of-week (dow) derived variable. Panels (b) and (c) show the seasonality of the average daily movement count, marked by the red vertical lines, and the subsequent reduction in mobility evident in the emergency period shown in panel (c). Panel (d) shows the rate of out-of-data coordinate values during the prediction period.}
  \Description{Exploratory data analysis of the mobility data important for the intuition and decisions in designing the model.}
  \label{fig:daily-trend}
\end{figure*}

Our analysis and the results of the competition demonstrate that the proposed GPT-based model exhibits promising performance in predicting human mobility, reinforcing its potential for applications across various domains. Furthermore, our investigation into the sensitivity of evaluation metrics towards generative parameters of the models adds a nuanced perspective to the assessment of human mobility prediction models. In essence, this paper serves as a stepping stone toward more accurate and robust human mobility prediction methods, with far-reaching implications for healthcare, urban planning, and beyond.

\section{Data and Metrics}

\subsection{Data}

The data used in this paper comes from an anonymized set of data on human mobility in an undisclosed metropolitan area in Japan. Spatiotemporal anonymization was conducted to ensure the privacy of individuals in the data. Two sets of data were released corresponding to the two tasks in the challenge. A more comprehensive description of the data is detailed in \citeN{yabe2023metropolitan}.

\subsubsection{Task 1 data.} The task 1 of the challenge requires a model that can predict human mobility in a "business-as-usual" period. Mobility trajectories from 100,000 individuals comprise the task 1 dataset. This dataset is collected over a 75-day period. The dataset contains full mobility trajectories for 80,000 individuals, while the remaining individuals only include full mobility information for 60 days. The remaining 15 days are undisclosed, representing the subset to be predicted by the model.

\subsubsection{Task 2 data.} The task 2 of the challenge demands a model that can predict human mobility in an "emergency" period that covers unusual human behaviour. Mobility trajectories from 25,000 individuals comprise the task 2 dataset. Similar to the task 1, the dataset is collected over a 75-day period. The dataset contains full mobility trajectories for 22,500 individuals, while the remaining individuals only include full mobility information for 60 days. Again, the remaining 15 days are undisclosed, representing the subset to be predicted by the model.


\subsubsection{Validation and Test split} We sampled 2000 users from each dataset to serve as the validation and test set, each having 1000 users. Individuals that were included in either the validation or test set will only have trajectories until the 60$^{th}$ day in the training data. The validation set is used to find the optimal checkpoint of the trained generative model. The test set was used primarily for assessing the model’s predictive performance and selecting the optimal generative parameters.

\subsection{Metrics}

The performance of the models in the challenge is evaluated using two metrics: the Dynamic Time Warping (DTW) and GEO-BLEU score. The DTW is a distance measure commonly used to measure the difference between time series data that may have different sequence lengths due to varying rates of observations \cite{1163055, vintsyuk1968speech}. The GEO-BLEU metric is inspired by the BLEU metric commonly used in the natural language processing literature \cite{shimizu2022geo}. A lower DTW score indicates better performance, whereas a higher GEO-BLEU score corresponds to a more performant model.


\section{Model and Methods}

In this section, we provide details on the analysis performed, the model proposed, and the strategies we implemented to transform the data, train the model, and generate predictions from the model.

\subsection{Analysis}

It is imperative to understand insights from the data to guide and justify modeling decisions. We performed simple exploratory analysis of the available data to gain some intuition that could be useful for implementing the model. We focused on exploring temporal insights from the data. The analysis of the global properties of the data reveals expected seasonality in mobility. While the actual times have been obfuscated, we can infer the occurrence of nighttime and daytime by plotting the distribution of observations by event stamp (Figure \ref{fig:daily-trend}, panels b and c). The evidence of periodicity in the data was a significant insight that drove one of the key modeling decisions, particularly simplifying the model by only learning one-week periods of the data. The drastic shift in mobility intensity during the emergency period seen in the latter part of panel (c) provided an insight into the decreased mobility of individuals. This change in behavior hinted at the likely impropriety of training the task 2 model with the full timeseries in the dataset, i.e., including the normal period in the training despite needing to predict only for the emergency period.

One additional insight we explored was the distribution of location visited by individuals after the 60-day period—the prediction period. This insight is essential because it can help guide whether we can fully rely on past trajectory for the inference or not. We found that in the prediction period, some of the coordinates, (x, y) tuples, have never appeared in the training period. Panel (d) in Fig. \ref{fig:daily-trend} shows the distribution of the rate of the out-of-training positions x, y, and their combination (x, y). While treating the x and y values of the location independently, we find that about 20\% of the x- or y-coordinates have not been seen in the past. Considering exact location coordinate (x, y), we see that about 40\% of locations visited in the prediction period by individuals have not been visited in the past. This is a crucial insight since it provides a significant bound to a model that relies solely on past visited locations.

\subsection{The GeoFormer Model}

Our proposed solution aims to reformulate human mobility as an abstracted sequence. This reformulation and abstraction allow us to apply models that can learn and generate sequential data. In particular, we establish an analogy between human mobility sequence and sequence of words in a sentence. This abstraction allows us to exploit all the deep learning machinery used to model and generate sentences, e.g., autoregressive decoder-only transformer models.

In the following, we provide the main details of the model we used to predict the mobility of users in both tasks. We first define the base model—we call the GeoFormer—and then discuss how we represented the data to fit the form required by the model.

\subsubsection{Generative Pre-trained Transformer (GPT)}

The GPT model is a transformer-based deep learning architecture for autoregressive modeling. The GPT uses a decoder transformer that takes in a sequence of tokens. The autoregressive learning is made possible by introducing masking with a training task designed to predict the next token in the sequence. This allows a trained GPT model to generate sequences. Theoretically, the GPT architecture models the conditional probability of generating a token $x$ at position $j$ given the past sequence $[x_0, x_1, x_2, …, x_{j-1}]$. Evidence for the impeccable ability of GPT architecture in modeling sequential data abounds \cite{radford_language_2019, padhi_tabular_2021, solatorio2023realtabformer, solatoriogenerating}. This makes the GPT formulation fully compatible with the problem at hand.

Specifically, we define the GeoFormer as a GPT model that learns the conditional probability distribution defined below,

\begin{equation}
\label{eq:cond-prob}
  x_{ij} \sim P(X|x_{i1}, x_{i2}, ..., x_{ij-1})
\end{equation}

where $i$ corresponds to the $i^{th}$ individual, $j$ corresponding to the $j^{th}$ time period, and $x_{ij}$ corresponding to the coordinate of the user $i$ at time $j$.

\subsubsection{Input linearization}

Reformulating the mobility problem as an NLP problem of sentence generation requires transforming the mobility data to fit into the chosen framework. We call this process \textit{input linearization}. We represent the location data as tokens in a sequence. We use the fact that one whole day of data is discretized into 30-minute intervals. This means that a full day will have at most 48 timesteps. We represent the daily trajectory for an individual using the full 48 timesteps despite the provided data only containing timesteps with observed locations, Fig. \ref{fig:daily-trend} - panel (a). So, we explicitly assign a special "empty" (\textbf{N}) token to timesteps without observations.

Representing the coordinates strictly as the tuple (x, y) is the precise approach in encoding the location information of individuals. However, this representation is suboptimal due to the existence of 500 distinct cells for both x and y coordinate values. Fully defining the geographic representation would require 250,000 unique location tokens. To mitigate this "explosion" of the token space, we independently represent the location of an individual. This means that we have 500 tokens for the x coordinates and 500 tokens for y coordinates. We distinguish these tokens as \textbf{\textrm{x<pos\_id>}} and \textbf{\textrm{y<pos\_id>}}, for x and y tokens, respectively shown in Annex \ref{annex:vocab}. That is, a coordinate is composed of two subsequent tokens: an x token followed by a y token. This choice is also influenced by the insight we earlier uncovered regarding the significant rates of out-of-training values for location (x, y) during the prediction period, as shown in Figure \ref{fig:daily-trend} - panel (d). That is, using (x, y) jointly will result in lower generalizability. This representation is a reasonable trade-off since the GPT model can learn the conditional probability of tokens as well.

The choice of the representation and the linearization of the data were largely influenced by the insights obtained from the analysis above. The seasonality observed in the data was considered in designing the model input. In particular, we represent the training data as a sequence of an 8-day mobility signature. Since the model is autoregressive, this 8-day mobility signature will allow us to generate the 8$^{th}$ day trajectory when we have input from the previous 7 days. This choice was made because of the clear seasonality in the one-week (7-day) period as shown in Figure \ref{fig:daily-trend}, panels b and c. However, one important limitation of the approach is worth noting which is the assumption that a one-week segment of mobility data sufficiently models the subsequent day.




While there is no explicit long-term memory for a trajectory of the individual beyond one week, the linearized input is designed to condition the model at the individual level. In particular, we prefix the individual's mobility data with a representation for the individual in the form of user id tokens. The learning algorithm is assumed to encode through the user id tokens the general long-term characteristics of the mobility specific to an individual. This is useful for predicting and generating the mobility trajectory of the individuals beyond the training data. An example of the fully linearized input is shown in Annex \ref{annex:linearized-input}. The full representation outlined complies with the input for the autoregressive framework, allowing us to generate subsequent mobility information of the individuals.







\subsection{Model Configuration and Training}

The model consists of 12 transformer layers having 24 attention heads, and 768 embedding dimensions with a 10\% dropout rate. The optimizer used is AdamW with beta values of (0.9, 0.999) and epsilon equal to $1^{-5}$. The learning rate scheduler follows a cosine curve with a maximum of $5^{-4}$ and a linear warm-up for 20,000 steps. A maximum gradient normalization of 5 was set.

The model for task 1 was trained for 5 epochs using all available data. For task 2, we fine-tuned the checkpoint corresponding to the task 1 model with the best validation metrics. While the nature of the two tasks is different, i.e., task 2 represents an emergency period, we found that performing fine-tuning of the task 1 trained model works reasonably well for task 2 as well. However, we limited the dataset to fine-tune the model for task 2 only on data from the 60$^{th}$ day until the 75$^{th}$ since the data distribution is different prior to the prediction period as depicted in Fig. \ref{fig:daily-trend} - panel (c).

\subsection{Generating predictions}

Prediction in the context of the GeoFormer is similar to the generative process performed in standard text applications of GPT. The process is autoregressive, meaning that each token in the sequence is generated one token at a time, and previously generated tokens are used to generate the next. A conditional generation is possible with the appropriate input data design.


\paragraph{The inference signature.}  To help the model generate the prediction, we exploit the provided signature in the data to be predicted. The data already specifies the time periods for which coordinates are to be predicted. So, we generate an expected input pattern from the data and only require the model to fill the values for the needed times. The signature shown in Annex \ref{annex:target-signature} indicates the values to be filled by the model as \textrm{x,y} while skipping predictions for times represented by \textrm{N}.

\paragraph{Limiting the candidate tokens.} Despite the insight that about 20\% of the x and y tokens in the prediction period have not appeared in the training period, we chose to limit the candidate tokens in the generation to those that have been already part of the past trajectories of the individual. We constrained the tokens specific to the day-of-week and the specific timestamp, with a window of 2 timestamps before and after. The window is used to account for the stochasticity in data collection, which could associate a location across neighboring timestamps due to issues in connectivity and other factors. This means that if we want a prediction for 6 a.m. on a Saturday, we only consider all the x and y locations previously visited by the individual at 5:00 a.m., 5:30 a.m., 6:00 a.m., 6:30 a.m., and 7:00 a.m. on previous Saturdays. Constraining the candidate tokens mitigates hallucinations by the model in generating locations that are too far from the individual's likely trajectory.






\paragraph{Generation parameters.} We experimented with some parameters of the generative algorithm in generating the trajectories. The temperature and the top-k parameters were the most useful based on our assessment. The top-p parameter was also varied, but it appears to produce similar effect with the top-k parameter.

\section{Results and Discussions}

We tracked the metrics on the test data to find the optimal set of parameters for the generation of the predictions. Our experiments suggest an inverse relationship between the GEO-BLEU and the DTW metrics as we vary the temperature parameter. As the temperature parameter approaches 1, the probability distribution of the tokens becomes more unbiased. In this regime, the GEO-BLEU score tends to improve. Decreasing the temperature parameter results in better DTW score, but negatively affects the GEO-BLEU score. Therefore, optimizing for both metrics requires careful tuning of the temperature parameter.

Another parameter we tuned was the top-k parameter. This parameter limits the tokens to be considered for generation only to tokens with the k highest probability. We varied this parameter and found that k = 5 produces generally better predictions measured by both metrics.

Part of the competition was an intermediate assessment of predictions. In this period, we submitted a version of the GeoFormer predictions for task 1 using only 6 transformer layers. The GeoFormer scored 0.3037 on the GEO-BLEU and scored 29.07 for the DTW on the final test data.

\begin{table}
  \centering
  \caption{Metrics values for the validation, test, and final data across tasks}
  \label{tab:metrics}
  \begin{tabular}{clccc}
    \toprule
    & \textbf{Metrics} & \textbf{Validation} & \textbf{Test} & \cellcolor{gray!25}\textbf{Final} \\
    \midrule
    \multirow{2}{*}{\textbf{Task 1}} & GEO-BLEU & 0.3047 & 0.3114 & \cellcolor{gray!25}0.3160 \\
                            & DTW      & 29.6978 & 29.8037 & \cellcolor{gray!25}26.2161 \\
    \midrule
    \multirow{2}{*}{\textbf{Task 2}} & GEO-BLEU & 0.2004 & 0.2053 & \cellcolor{gray!25}0.1828 \\
                            & DTW      & 38.0069 & 43.4332 & \cellcolor{gray!25}37.7815 \\
    \bottomrule
  \end{tabular}
\end{table}


Summary of the validation, test, and final scores for the models we selected for submission is reported in Table \ref{tab:metrics}. The scores show a considerable variation in the validation and test DTW scores compared with the scores measured on the final assessment groups, versus the GEO-BLEU. The model appears to be more stable when applied to the task 1 compared with the task 2 dataset. This may be due to the larger task 1 dataset.

\section{Conclusion}

In this paper, we detailed a generative deep learning model for predicting human mobility data. The GeoFormer model achieved a top score in the challenge having optimal scores for both the GEO-BLEU and the DTW metrics across the two types of mobility data tested. We believe that the success of the GeoFormer in the HuMob Challenge 2023 will pave the way for more applications of generative deep learning models in solving problems related to human mobility.

\bibliographystyle{ACM-Reference-Format}
\bibliography{geoformer}

\appendix

\section{Data processing}

\subsection{Vocabulary}
\label{annex:vocab}

The linearized trajectory of an individual consists of tokens from the following set. These tokens are mapped to learnable embeddings of the model. The day-of-week tokens are expected to learn the variations across different days in a week. The x and y coordinates are independently represented instead of creating unique tokens for a tuple (x, y). This helps reduce the number of tokens in the model, and also help the model generalize.

\begin{minted}[
frame=lines,
framesep=2mm,
bgcolor=brickred!5,
baselinestretch=1.2,
fontsize=\scriptsize,
xleftmargin=10pt,
numbersep=5pt,
linenos
]{python}
# Special tokens:
    <eos>, <|data|>, <|sep|>

# The day-of-week tokens:
    <|dow0|>, <|dow1|>, <|dow2|>, <|dow3|>, <|dow4|>, <|dow5|>, <|dow6|>

# User id tokens
    0, 1, 2, 3, 4, 5, 6, 7, 8, 9

# Location tokens:
    N
    x000, x001, x002, x003, x004, ..., x499
    y000, y001, y002, y003, y004, ..., y499
\end{minted}

\subsection{Example linearized input}
\label{annex:linearized-input}

This signature represents a trajectory for user 71000. The data starts on day-of-week 6. This will be an input to the trained model to generate the trajectory for the next day. The special token \textrm{<|sep|>} conditions the start of the prediction.


\begin{minted}[
frame=lines,
framesep=2mm,
bgcolor=brickred!5,
baselinestretch=1.2,
fontsize=\scriptsize,
xleftmargin=10pt,
numbersep=5pt,
linenos
]{python}
71000
<|data|>
<|dow6|>NNNNNNNNNNNNNNNNNNNx129y088x129y090x128y086x128y087x131y092N
    x132y089x132y091x132y092Nx128y089x126y091NNNx131y092x132y088NN
    x132y087NNx126y086NNNNNN
<|dow0|>NNNNNNNNNNNNNNNx135y076x126y084x127y085x133y078x126y086NN
    x128y092x126y085x126y086x127y086NNx127y092x127y089x131y092x131y090
    x126y088x124y092x127y085Nx127y085NNNx130y086x130y090NNx126y086NNN
<|dow1|>NNNNNNNNNNNNNNNNNx126y087x127y088x127y085x124y093x125y091x125y084
    x126y087x131y091NNNNNNNNNx128y089x126y087x126y086x128y087x126y088
    x131y090Nx132y093x132y087x131y080x126y086NNN
<|dow2|>NNNNNNNNNNNNNNNNNx128y089Nx130y087x131y091NNNx128y089x127y086N
    x127y096x128y101x126y097Nx131y092Nx118y071Nx120y077x126y086x128y087
    x126y086Nx131y090NNNNNNx125y088
<|dow3|>NNNNNNNNNNNNNNNNx144y072x149y075NNx149y075NNx161y062x141y075
    x131y092x126y086x126y089x125y100x121y102NNx121y101x122y102Nx126y100
    x127y089x127y086NNNNNNNNNN
<|dow4|>NNNNNNNNNNNNNNNNNx126y082x141y079x154y074x157y073NNNNNx157y082
    x130y087NNx131y092x125y088Nx126y086x128y099x130y094NNNNx142y079NNN
    x131y092NNN
<|dow5|>x128y089x126y086NNx126y086NNNNNNNNNNNx131y090NNx128y089x129y089
    x131y092NNx132y092NNNNNNNNNNNx131y092NNNNNNx126y086x131y092NNN
<|sep|>
\end{minted}

\subsection{Example target signature}
\label{annex:target-signature}
The target signature is used to inform the generative algorithm to limit the scope of prediction only on periods where location values are expected. The N represents time periods where no data is expected, the xy represents the generation of a sequence of coordinates (x, y) in the given period. For example, when the prediction corresponds to an \textrm{x}, only location tokens for the x coordinate are considered in the generation. The leading number 6 represents the day-of-week, which will be parsed, helping contextualize the generation of the coordinates.


\begin{minted}[
frame=lines,
framesep=2mm,
bgcolor=brickred!5,
baselinestretch=1.2,
fontsize=\scriptsize,
xleftmargin=10pt,
numbersep=5pt,
linenos
]{python}
6NNNNxyNNNNNNNNNNNNNNxyNxyxyxyNNNxyxyNxyxyNxyxyxyNNNNxyNNxyNNNN
\end{minted}

\subsection{Evaluation loss traces}

\begin{figure}[h]
  \centering
  \includegraphics[width=\linewidth]{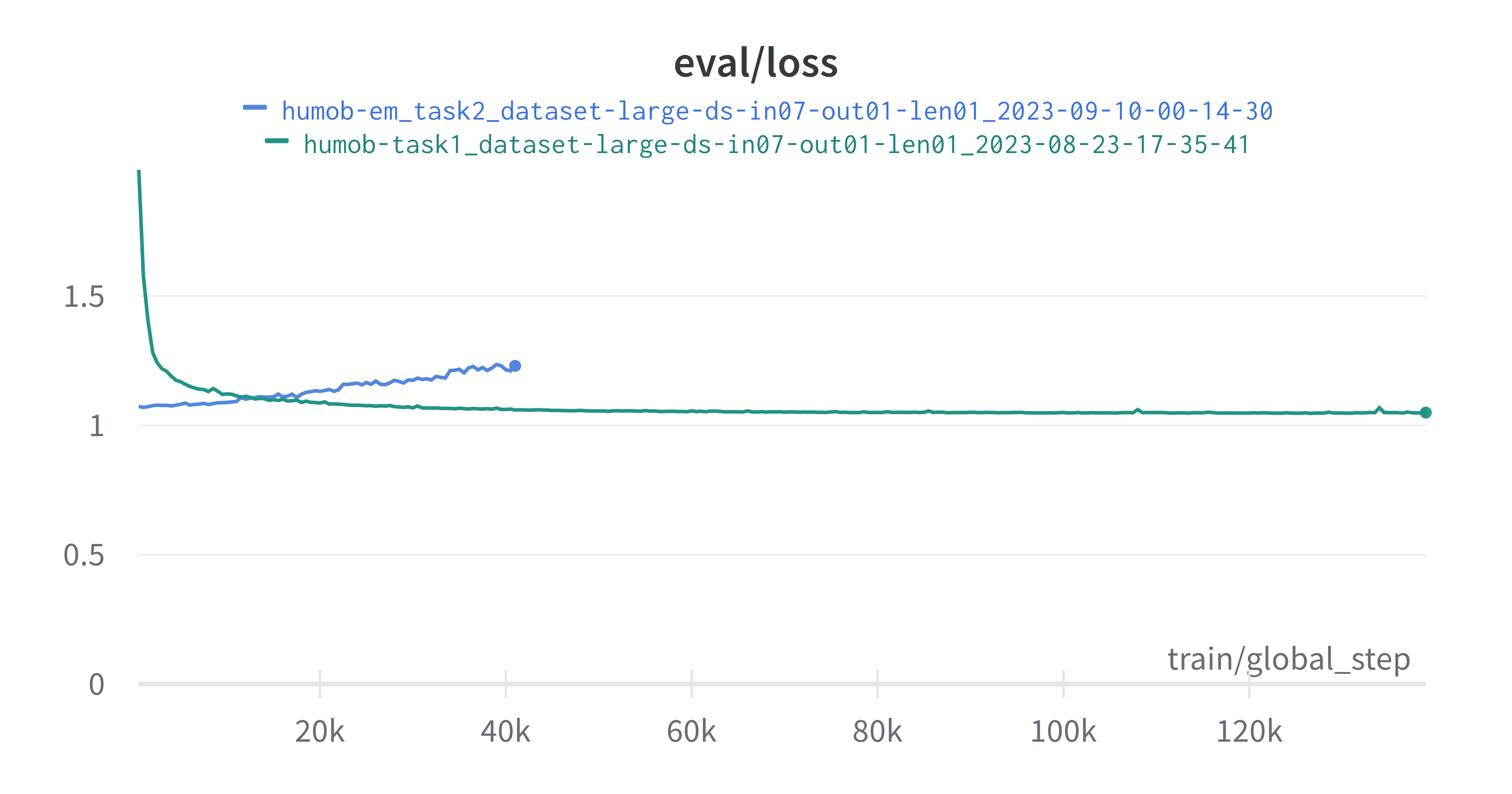}
  \caption{Evaluation loss traces for the models. The evaluation loss for the task 1 dataset does not diverge despite training relatively long. On the other hand, the fine-tuned model for task 2 briefly decreased but eventually diverged. We chose the model for task 2 with the lowest validation metric.}
\end{figure}

\end{document}